# Combination of PCA with SMOTE Resampling to Boost the Prediction Rate in Lung Cancer Dataset

Mehdi Naseriparsa
Islamic Azad University
Tehran North Branch
Dept. of Electrical & Computer Engineering
Tehran, Iran

Mohammad Mansour Riahi Kashani
Islamic Azad University
Tehran North Branch
Dept. of Electrical & Computer Engineering
Tehran, Iran

## ABSTRACT
Classification algorithms are unable to make reliable models on the datasets with huge sizes. These datasets contain many irrelevant and redundant features that mislead the classifiers. Furthermore, many huge datasets have imbalanced class distribution which leads to bias over majority class in the classification process. In this paper combination of unsupervised dimensionality reduction methods with resampling is proposed and the results are tested on Lung-Cancer dataset. In the first step PCA is applied on Lung-Cancer dataset to compact the dataset and eliminate irrelevant features and in the second step SMOTE resampling is carried out to balance the class distribution and increase the variety of sample domain. Finally, Naïve Bayes classifier is applied on the resulting dataset and the results are compared and evaluation metrics are calculated. The experiments show the effectiveness of the proposed method across four evaluation metrics: Overall accuracy, False Positive Rate, Precision, Recall.

## 1. INTRODUCTION
Data mining and machine learning depend on classification which is the most essential and important task. Many experiments are performed on medical datasets using multiple classifiers and feature selection techniques. The growth of the size of data and number of existing databases exceeds the ability of humans to analyze this data, which creates both a need and an opportunity to extract knowledge from databases [1]. Extracting huge amount of data to find a reasonable pattern has become a heated issue in the realm of data mining.

Assareh[2] proposed a hybrid random model for classification that uses the subspace and domain of the samples to increase the diversity in the classification process. In another attempt Duangsoithong and Windeatt [3] presented a method for reducing dimensionality in the datasets which have huge amount of attributes and few samples. Dhiraj[4] used clustering and K-mean algorithm to show the efficiency of this method on huge amount of data. Jiang[5] proposed a hybrid feature selection algorithm that takes the benefit of symmetrical uncertainty and genetic algorithms. Zhou[6] presented a new approach for classification of multi class data. The algorithm performed well on two kinds of cancers. Naseriparsa and Bidgoli[7] proposed a hybrid feature selection method to improve performance of a group of classification algorithms.

In this paper, in section 2,3,4,5 and 6 we focus on the definition of feature extraction, PCA, Naïve Bayes, SMOTE and present some information about Lung-Cancer dataset. In section 7 evaluation metrics for performance evaluation is presented. In section 8, the proposed combined method is described. In section 9 performance evaluation results are presented and in section 10 the results are interpreted. Conclusions are given in section 11.

## 2. FEATURE EXTRACTION
Today we face huge datasets that contain some million samples and thousands of features. Preprocessing methods extract features in order to reduce dimensionality and eliminate irrelevant and redundant data and prepare data for data mining analysis. Feature selection methods try to find a subset of features from the main dataset. Furthermore, feature selection methods lead to generation of new feature set that is based on the main dataset.

Data preprocessing includes conversion of the main dataset to a new dataset and simultaneously reducing dimensionality by extracting the most suitable features. Conversion and dimensionality reduction will result in a better understanding of the existing patterns in the dataset and more reliable classification by observing the most important data which keeps the maximum properties of the main data. This approach results in better generalization in the classifiers. The dimensionality reduction leads to the conversion of n-dimensional dataset to m-dimensional one in which this relation exists: (m <=n). Furthermore, this conversion can be expressed by a linear conversion function shown in equation 1.

$$Y = F(X) \qquad (1)$$

In equation 1, X shows the main dataset patterns with n-dimensional space. Y shows the converted patterns with m-dimensional space. Some benefits for conversion and dimensionality reduction are:

- Redundancy removal.
- Dataset compression.
- Efficient feature selection which help design a data model and a better understanding of generated patterns
- High dimensional data visualization on low dimensional space for data mining [8].

## 3. PRINCIPAL COMPONENT ANALYSIS

In many problems, the measured data vectors are high-dimensional but we may have reason to believe that the data lie near a lower-dimensional manifold. In other words, we may believe that high-dimensional data are multiple, indirect measurements of an underlying source, which typically cannot be directly measured. Learning a suitable low-dimensional manifold from high-dimensional data is essentially the same as learning this underlying source. Principal components analysis (PCA) [9] is a classical method that provides a sequence of best linear approximations to a given high-dimensional observation. PCA is a way of identifying patterns in data, and expressing the data in such a way as to highlight their similarities and differences. Since patterns in data can be hard to find in data of high dimension, where the luxury of graphical representation is not available, PCA is a powerful tool for analyzing data. The other main advantage of PCA is that once you have found these patterns in the data, and you compress the data, by reducing the number of dimensions, without much loss of information.

## 4. NAÏVE BAYES CLASSIFIER

The Naive Bayes algorithm is based on conditional probabilities. It uses Bayes' Theorem, a formula that calculates a probability by counting the frequency of values and combinations of values in the historical data. Bayes' Theorem finds the probability of an event occurring given the probability of another event that has already occurred. If B represents the dependent event and A represents the prior event, Bayes' theorem can be stated as follows. Bayes' Theorem is shown in equation 2.

$$Prob(B\ given\ A) = \frac{Prob(A\ and\ B)}{Prob(A)} \quad (2)$$

In equation 2, to calculate the probability of B given A, the algorithm counts the number of cases where A and B occur together and divides it by the number of cases where A occurs alone.

Naive Bayes makes the assumption that each predictor is conditionally independent of the others. For a given target value, the distribution of each predictor is independent of the other predictors. In practice, this assumption of independence, even when violated, does not degrade the model's predictive accuracy significantly, and makes the difference between a fast, computationally feasible algorithm and an intractable one.

The Naive Bayes algorithm affords fast, highly scalable model building and scoring. It scales linearly with the number of predictors and rows. The build process for Naive Bayes is parallelized. (Scoring can be parallelized irrespective of the algorithm). Naive Bayes can be used for both binary and multiclass classification problems and deals well with missing values [10].

## 5. SYNTHETIC MINORITY OVERSAMPLING TECHNIQUE

Often real world datasets are predominantly composed of normal examples with only a small percentage of abnormal or interesting examples. It is also the case that the cost of misclassifying an abnormal example as a normal example is often much higher than the cost of the reverse error. Under sampling of the majority (normal) class has been proposed as a good means of increasing the sensitivity of a classifier to the minority class. By combination of over-sampling the minority (abnormal) class and under-sampling the majority (normal) class, the classifiers can achieve better performance than only under-sampling the majority class. SMOTE adopts an over-sampling approach in which the minority class is over-sampled by creating synthetic examples rather than by over-sampling with replacement. SMOTE can control the number of examples and distribution to achieve the purpose of balancing the dataset through synthetic new examples, and it is an effective over-sampling method to solve the over-fitting problem because of decision interval is too narrow. The synthetic examples are generated in a less application specific manner, by operating in feature space rather than sample domain. The minority class is over-sampled by taking each minority class sample and introducing synthetic examples along the line segments joining any of the k minority class nearest neighbors. Depending upon the amount of over-sampling required, neighbors from the k nearest neighbors are randomly chosen [11].

## 6. LUNG CANCER DATASET

To evaluate the performance of the proposed method, Lung-Cancer dataset from UCI Repository of Machine Learning databases [12] is selected and Naïve Bayes classification algorithm is applied on five different datasets which are obtained from application of five methods. Lung-Cancer dataset contains 56 features and 32 samples which is classified into three groups. The data described 3 types of pathological lung cancers (Type A, Type B, Type C).

## 7. EVALUATION METRICS

A classifier is evaluated by a confusion matrix as illustrated in table1. The columns show the predicted class and the rows show the actual class [13]. In the confusion matrix, TN is the number of negative samples correctly classified (True Negatives), FP is the number of negative samples incorrectly classified as positive(False Positives), FN is the number of positive samples incorrectly classified as negative(False Negatives) and TP is the number of positive samples correctly classified (True Positives).

**Table1. Confusion Matrix**

|  | Predicted Negative | Predicted Positive |
|---|---|---|
| Actual Negative | TN | FP |
| Actual Positive | FN | TP |

Overall accuracy is defined in equation 3.

$$Accuracy = \frac{(TP + TN)}{(TP + FP + TN + FN)} \quad (3)$$

Overall accuracy is not an appropriate parameter for performance evaluation when the data is imbalanced. The nature of some problems require a fairly high rate of correct detection in the minority class and allows for a small error rate in the majority class while simple overall accuracy is clearly not appropriate in such cases[14]. Actually, overall accuracy is biased over the majority class which contains more samples and this measure does not represent the minority class accuracy.

From the confusion matrix in table1, the expressions for FP rate, Recall and Precision are derived which are presented in equations 4, 5 and 6.

$$\text{FP Rate} = \frac{FP}{(TN + FP)} \quad (4)$$

$$\text{TP Rate} = Recall = \frac{TP}{(TP + FN)} \quad (5)$$

$$\text{Precision} = \frac{TP}{(TP + FP)} \quad (6)$$

The main goal for learning from imbalanced datasets is to improve the recall without hurting the precision. However, recall and precision goals can be often conflicting, since when increasing the true positive for the minority class, the number of false positives can also be increased; this will reduce the precision.

## 8. COMBINATION OF PCA WITH SMOTE RESAMPLING

### 8.1 First Step (PCA Application)
For the first step, PCA is applied on Lung-Cancer dataset. PCA compacts the dataset feature space from 56 features to 18 features. Compacted dataset cover 90 percent of the main dataset variance. This amount of reduction is considerable in the feature space and paves the way for the next step in which the sample domain distribution is targeted.

### 8.2 Second Step (SMOTE Application)
For the second step, SMOTE resampling method is carried out on the dataset. Lung-Cancer dataset has 3 classes (class A, class B, class C). Hence, class A with 9 samples is considered as the minority class in the first run of SMOTE. After the first run of SMOTE, the resulting dataset contains 18 samples for class A, 13 samples for class B and 10 samples for class C.

In the second run of SMOTE, class C with 10 samples is considered as the minority class. After the second run of SMOTE sample domain contains 18 samples for class A, 13 samples for class B and 18 samples for class C.

In the third run of SMOTE, class B with 13 samples is considered as the minority class. After the third run of SMOTE, the resulting dataset contains 18 samples for each class in the dataset.

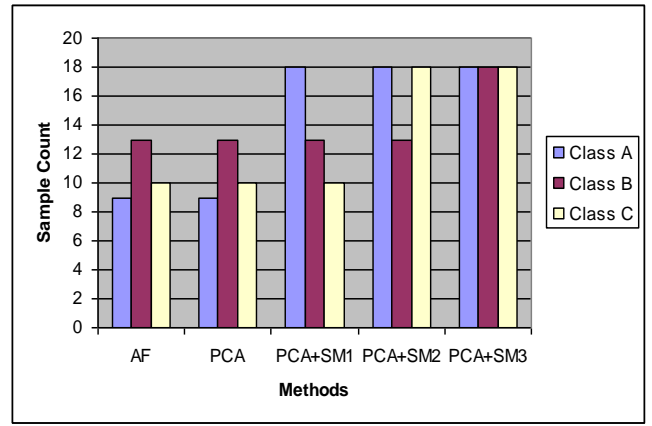

**Figure 1: class distribution for different methods**

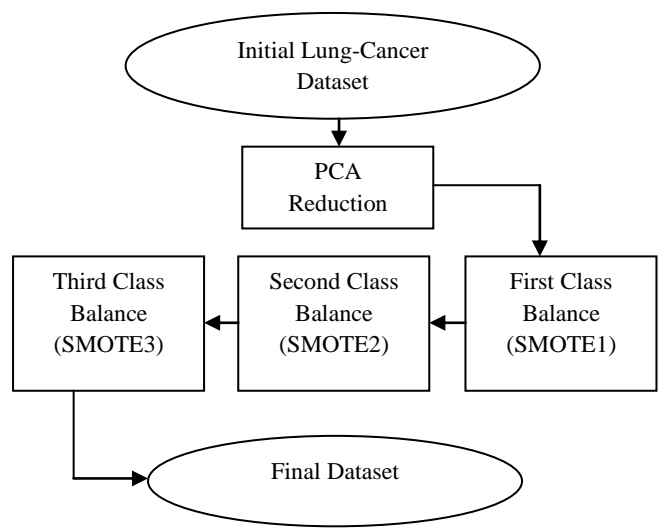

**Figure 2: Flow diagram of different steps in the proposed method**

## 9. PERFORMANCE EVALUATION
### 9.1 Experiment Results
In table2, the results of the experiments in different steps are presented.

**Table 2. Results from running our proposed method on Lung Cancer dataset**

| Steps | Features | Samples |
|---|---|---|
| **Initial State** | 56 | 32 |
| **PCA Reduction** | 18 | 32 |
| **SMOTE1** | 18 | 41 |
| **SMOTE2** | 18 | 49 |
| **SMOTE3** | 18 | 54 |

## 9.2 Overall Accuracy

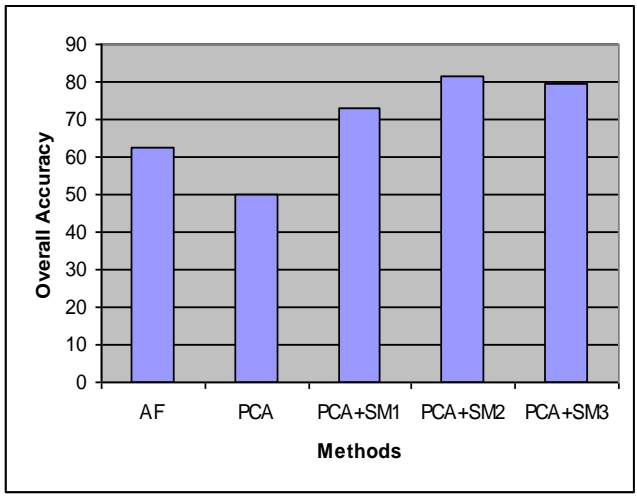

**Figure 3: Overall Accuracy parameter values for different methods**

In figure3, overall accuracy parameter is calculated for different methods. When all features are used for classification, the accuracy is above 60 percent. After the application of PCA, we observe that the accuracy has been reduced to 50 percent. This reduction is for reducing the feature space from 56 features to 18 features. Some of useful information has been removed during the application of PCA on the dataset because PCA is an unsupervised dimensionality reduction method and no search is carried out during the reduction phase on the feature space. The proposed method uses both PCA and resampling to use the benefits of both methods simultaneously. After running PCA, SMOTE resampling method is applied and the accuracy in this condition has been increased to above 70 percent. This increase is due to the use of SMOTE resampling right after the running of PCA. Actually, in this condition, SMOTE contributes to increase the variety of sample domain and compensate the loss of some information which is occurred in the application of PCA. In another attempt SMOTE is run on the resulting dataset to balance the distribution of the class C in the dataset. The accuracy in this time is above 80 percent which is the optimal situation. In the last try, SMOTE is run to balance the class B. in this step, we observe that the accuracy is reduced to below 80 percent which is lower comparing to the previous step. Hence, for overall accuracy parameter we reach the optimal situation after running the SMOTE two times and balancing the classes A and C.

## 9.3 False Positive Rate

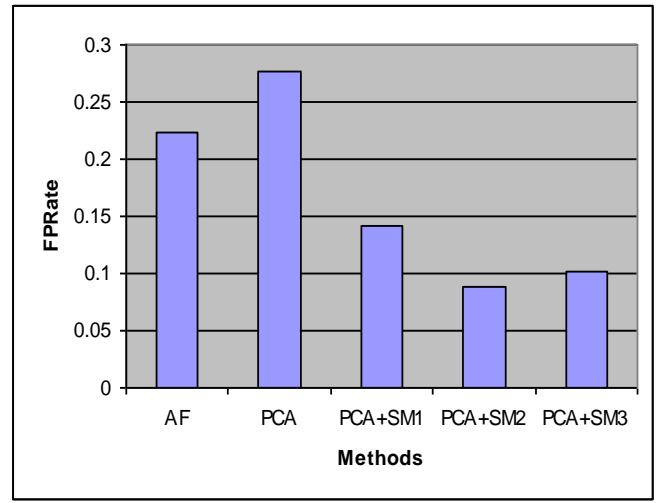

**Figure 4: False Positive Rate parameter values for different methods**

In figure4, False Positive Rate parameter is calculated for different methods. If False Positive Rate increases, we can interpret that the performance of classification is low because the number of incorrectly classified samples in the minority class is on the rise. Reversely, if False Positive Rate decreases, we can interpret that the performance of classification is high because the number of incorrectly classified samples for the minority class is on the decline. When all features are used for classification, this rate is above 0.2. After the application of PCA, we observe that this rate has been increased to 0.27. This increase shows that reducing the feature space from 56 features to 18 features has led to loss of some information during the application of PCA and this happening plays an important part in the increase of False Positive Rate. After the running of SMOTE for the first time, we observe that this rate decreases to below 0.15 which shows that using SMOTE after PCA helps to increase the performance of classification. After the application of SMOTE for the second time, the optimal result is obtained and False Positive Rate reaches the lowest value to below 0.1. After the third application of SMOTE on the dataset, False Positive Rate increases to over 0.1 which shows that the performance is reduced comparing to the previous step.

## 9.4 Precision

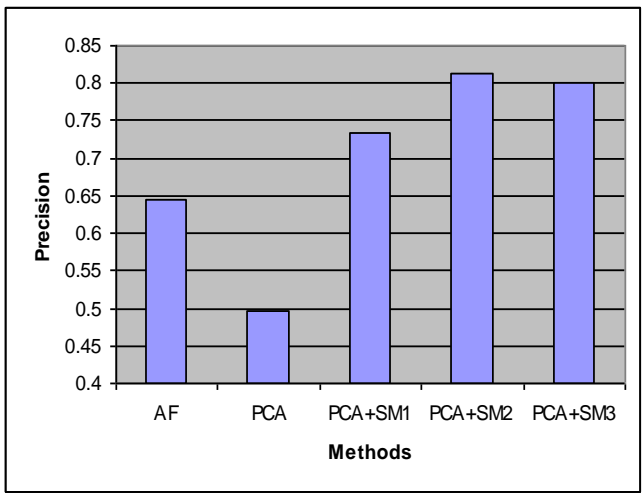

**Figure 5: Precision parameter values for different methods**

In figure5, Precision parameter is calculated for different methods. When all features are used for classification, the Precision is near 0.65. After the application of PCA, we observe that Precision has been reduced to below 0.5. After running PCA, SMOTE resampling method is applied and the Precision in this condition has been increased to 0.73. This increase is due to the use of SMOTE resampling right after the running of PCA. Actually, SMOTE contributes to increase the variety of sample domain and compensate the loss of some information which is occurred in the application of PCA. After the second application of SMOTE, Precision is increased to above 0.8 which is the optimal situation. After the third application of SMOTE on the dataset, Precision reduces to 0.8 which shows that the performance is reduced comparing to the previous step.

In fact, in the third run of SMOTE on Lung-Cancer dataset, Precision parameter is reduced from 0.813 to 0.8. This reduction is due to the generation of synthetic samples for class B which once has been the majority class in the initial state. These samples do not help to improve Precision parameter.

## 9.5 Recall

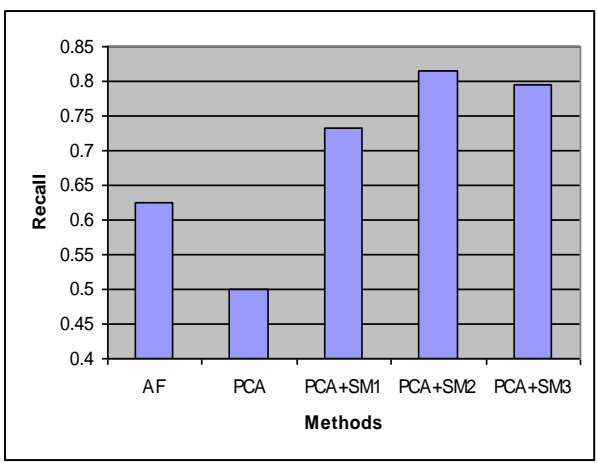

**Figure 6: Recall parameter values for different methods**

In figure6, Recall parameter is calculated for different methods. When all features are used for classification, this rate is above 0.6. After the application of PCA, we observe that this rate has been reduced to 0.5. This reduction shows that reducing the feature space from 56 features to 18 features has led to loss of some information during the application of PCA and this happening plays an important part in the reduction of Recall parameter. After the running of SMOTE for the first time, we observe that Recall increases to above 0.7 which shows that using SMOTE after PCA helps to increase the performance of classification. After the application of SMOTE for the second time, the optimal result is obtained and Recall reaches the highest value to above 0.8. After the third application of SMOTE on the dataset, Recall decreases to below 0.8 which shows that the performance is reduced comparing to the previous step.

## 10. ANALYSIS AND DISCUSSION

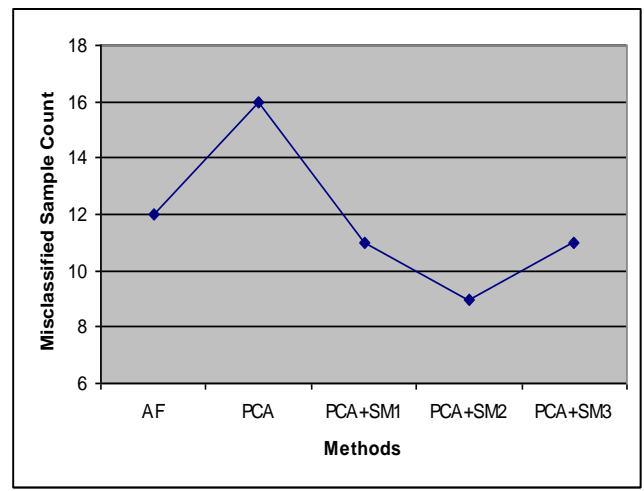

**Figure 7: Number of misclassified samples for different methods.**

From figure 7, we observe that when all features are used for classification, the number of incorrectly classified samples is 12. After running PCA on the dataset, the number of misclassified samples rises to 16. In fact, this increase is due to reduction of feature space during the application of PCA in which some information is removed. In the third method, SMOTE resampling is carried out and in this method the number of misclassified samples reduces to 11. After the second run of SMOTE, the number of misclassified samples reduces to its lowest value to 9. Finally, in the third run of SMOTE, the number of incorrectly classified samples rises to 11. This shows that the performance deteriorated comparing to the previous step.

As it is clear from the performance evaluation results, the best performance is obtained when PCA is used and SMOTE resampling is run on Lung Cancer dataset for two times. All parameters evaluated in this method, reach their peak value in comparison with other methods. In Lung-Cancer dataset, we have two minority classes (Class A- Class C). Hence, after the application of SMOTE for two times, the distribution of the two minority classes gets balanced. After the third run of SMOTE, synthetic samples are generated for class B which once has been the majority class in the previous steps and these samples do not contribute to boost the performance of classification. Therefore, the best method is using PCA with SMOTE which is run on the dataset for two times.

## 11. CONCLUSION

A combination of unsupervised dimensionality reduction with resampling is proposed to reduce the dimension of Lung-Cancer dataset and compensate this reduction with resampling. PCA is used to reduce the feature space and lower the complexity of classification. PCA try to keep the main characteristics of initial dataset in the compacted dataset; however, some useful information is lost during the PCA reduction. SMOTE resampling is used to work on the sample domain and increase the variety of sample domain and balance the distribution of classes in the dataset. Experiments and evaluation metrics show that performance improved while feature space reduced more than a half and this leads to lower the cost and complexity of classification process.